\documentclass[10pt,twocolumn,letterpaper]{article}

\usepackage[pagenumbers]{wacv}      % To produce the REVIEW version for the algorithms track
\usepackage{graphicx}
\usepackage{amsmath}
\usepackage{amssymb}
\usepackage{booktabs}
\usepackage{subcaption}
\usepackage[percent]{overpic}
\usepackage{float}
\usepackage{multirow}
\usepackage{color, colortbl}
\usepackage{xcolor}
\usepackage{makecell}
\usepackage{etoolbox}
\usepackage{fdsymbol}
\usepackage[flushleft]{threeparttable}
\usepackage[symbol]{footmisc}

\newcommand\tstrut{\rule{0pt}{2.4ex}}
\newcommand\bstrut{\rule[-1.0ex]{0pt}{0pt}}
\definecolor{LightGrey}{rgb}{0.9,0.9,0.9}

\definecolor{future}{HTML}{06D6A0}
\definecolor{present}{HTML}{118AFF}
\definecolor{past}{HTML}{E84545}
\definecolor{trajectory}{HTML}{808080}
\definecolor{ensemble}{HTML}{F07B3F}

\usepackage[pagebackref,breaklinks,colorlinks]{hyperref}

\usepackage[capitalize]{cleveref}
\crefname{section}{Sec.}{Secs.}
\Crefname{section}{Section}{Sections}
\Crefname{table}{Table}{Tables}
\crefname{table}{Tab.}{Tabs.}

\newcommand\SupplementaryMaterial{%
  \xdef\presupfigures{\arabic{figure}}% save the current figure number
  \xdef\presupsections{\arabic{section}}% save the current section number
  \renewcommand\thefigure{S\fpeval{\arabic{figure}-\presupfigures}}
  \renewcommand\thesection{S\fpeval{\arabic{section}-\presupsections}}
  \renewcommand{\thetable}{S\arabic{table}}
  \renewcommand{\theequation}{S\arabic{equation}}
}

\def\wacvPaperID{1254} % *** Enter the WACV Paper ID here
\def\confName{WACV}
\def\confYear{2024}

\begin{document}

\title{Holistic Representation Learning for Multitask Trajectory Anomaly Detection}

\author{Alexandros Stergiou \hspace{2em} Brent De Weerdt \hspace{2em} Nikos  Deligiannis\\
Vrije Universiteit Brussel, Belgium \& imec, Belgium \\
{\tt\small <first>.<last>@vub.be}
%\and
}
\maketitle

%%%%%%%%% ABSTRACT
\begin{abstract}
   Video anomaly detection deals with the recognition of abnormal events in videos. Apart from the visual signal, video anomaly detection has also been addressed with the use of skeleton sequences. We propose a holistic representation of skeleton trajectories to learn expected motions across segments at different times. Our approach uses multitask learning to reconstruct any continuous unobserved temporal segment of the trajectory allowing the extrapolation of past or future segments and the interpolation of in-between segments. We use an end-to-end attention-based encoder-decoder. We encode temporally occluded trajectories, jointly learn latent representations of the occluded segments, and reconstruct trajectories based on expected motions across different temporal segments. Extensive experiments on three trajectory-based video anomaly detection datasets show the advantages and effectiveness of our approach with state-of-the-art results on anomaly detection in skeleton trajectories\footnote[2]{Code is available at \href{https://alexandrosstergiou.github.io/project_pages/TrajREC/index.html}{alexandrosstergiou/TrajREC} }.
\end{abstract}

%%%%%%%%% BODY TEXT
\section{Introduction}
\label{sec:intro}

\begin{figure}[t]
    \centering
    \includegraphics[width=.98\linewidth]{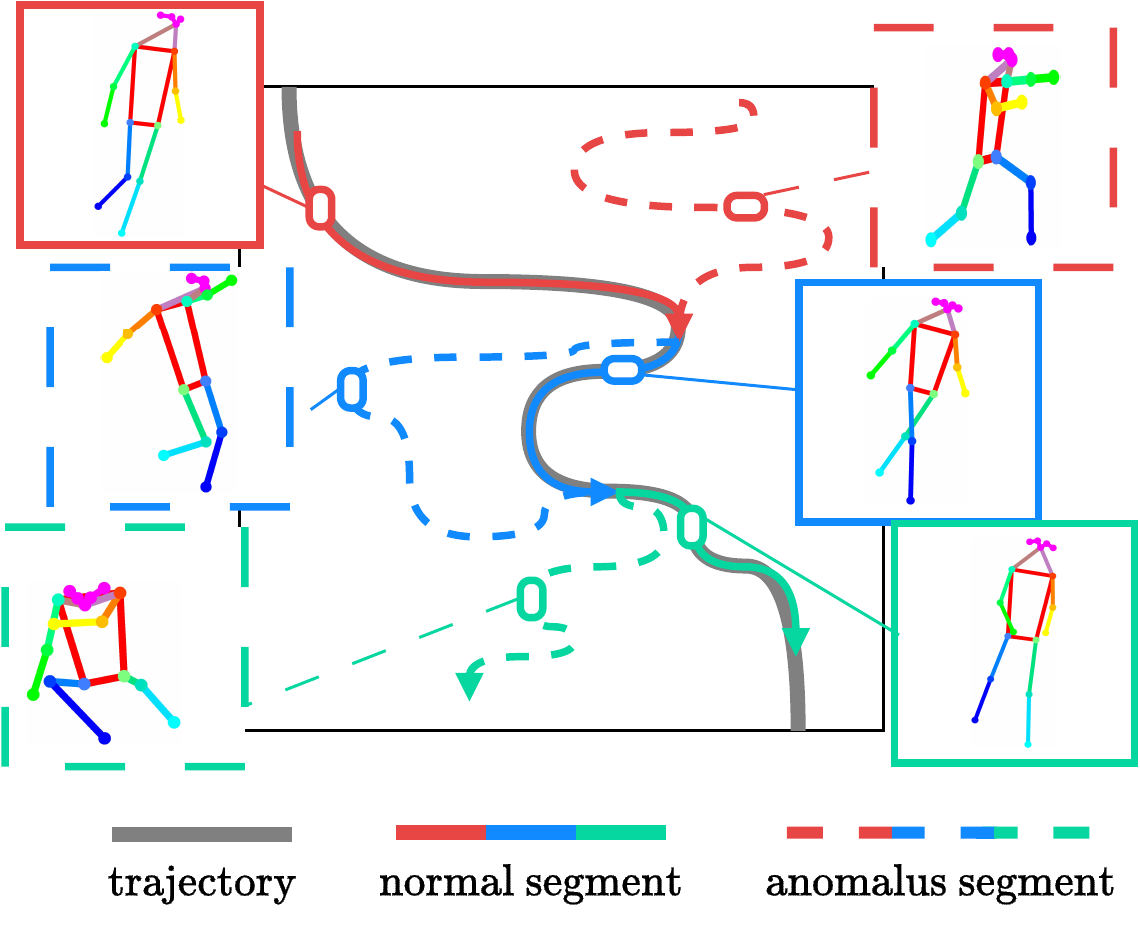}
    \caption{\textbf{Holistic representation of trajectory segements}. Given a normal trajectory in \textcolor{trajectory}{grey}, past segments (\textcolor{past}{Pst}) are extrapolated based on future segments, present segments (\textcolor{present}{Prs}) are interpolated from proceeding and succeeding segments, and future segments (\textcolor{future}{Ftr}) are extrapolated from past segments. During training, reconstructed trajectories are compared to ground truth trajectories. In inference, the reconstruction error of task-specific segments is used to distinguish normal and anomalous trajectories.}
    \label{fig:normal_anomalous_segments}
    \vspace{-1em}
\end{figure}

% Expected motions and their representation with trajectories 
Everyday life is full of expected behaviors. Examples include; walking towards a destination, sitting down, or riding a bicycle. Such actions and behaviors are defined by the predictability of their motions that are part of an expected sequence. In computer vision, motion can be represented by trajectories, segments of which include \emph{past}, \emph{present}, and \emph{future} events. Anomalies can occur at different times within a trajectory and are usually perceived through abrupt or sudden motion changes. The resulting anomalous trajectory segments deviate from the expected trajectory path, as illustrated in Figure~\ref{fig:normal_anomalous_segments}.

% Task definition and limitations of future extrapolation
Video anomaly detection (VAD) is the open-set task of detecting anomalous actions and motions by learning robust representations of expected behaviors in normal events. The detection of abnormal events in videos has gained traction with applications in surveillance~\cite{sultani2018real,nayak2021comprehensive}, social networks~\cite{sharma2018nhad}, and monitoring of daily activities~\cite{parvin2018anomaly}. The majority of trajectory-based VAD methods~\cite{markovitz2020graph,morais2019learning,flaborea2023contracting,flaborea2023multimodal} are trained to only extrapolate the future of normal trajectories. Abnormalities are detected at inference from the reconstruction error between predicted expected trajectories and the ground truth. Due to the continuous nature of trajectories, their reconstruction at any point in time is challenging in real-world scenarios as the future or past may not be observed and occlusions or pose estimation errors may lead to partial observations of the present.

% Overview of framework
We introduce a unifying framework that jointly addresses these challenges and enables the representation and reconstruction of expected normal motions at different trajectory segments. We use a holistic representation of skeleton trajectories as a composition of past, present, and future segments. We use an encoder-decoder~\cite{he2022masked} to first encode partial observations of normal trajectories with different temporal segments being occluded. Latent representations of the occluded segments are learned regressively through positive and negative pairs drawn from full trajectories. The learned representations are then decoded back to the input space and compared to ground truth segments. During training, our model learns to encapsulate information relating to normal motions and inter/extrapolates the occluded trajectory segments. At inference, the model predicts a corresponding normal trajectory that is compared to the ground truth trajectory of the unseen abnormal scene. Our method's simplicity allows the detection of abnormalities across multiple temporal locations within the trajectory and the use of trajectories with different temporal lengths.

% Summary of contributions
In summary, our contributions are as follows: (i) We propose a holistic representation of trajectories for VAD to detect anomalies in past, present, or future trajectory segments. (ii) We train an encoder-decoder model with multitask learning to jointly learn latent representations of occluded past, present, and future trajectory segments and reconstruct them back to the input space. (iii) We evaluate our approach on three skeleton VAD datasets: HR-ShanghaiTech Campus (HR-STC)~\cite{luo2017revisit,morais2019learning}, HR-Avenue~\cite{lu2013abnormal,morais2019learning}, and (HR-)UBnormal~\cite{acsintoae2022ubnormal,flaborea2023contracting}. We consistently outperform prior work on extrapolating the future and we additionally provide baselines on the remaining two tasks.

%------------------------------------------------------------------------

\begin{figure*}[t]
    \centering
    \includegraphics[width=\linewidth]{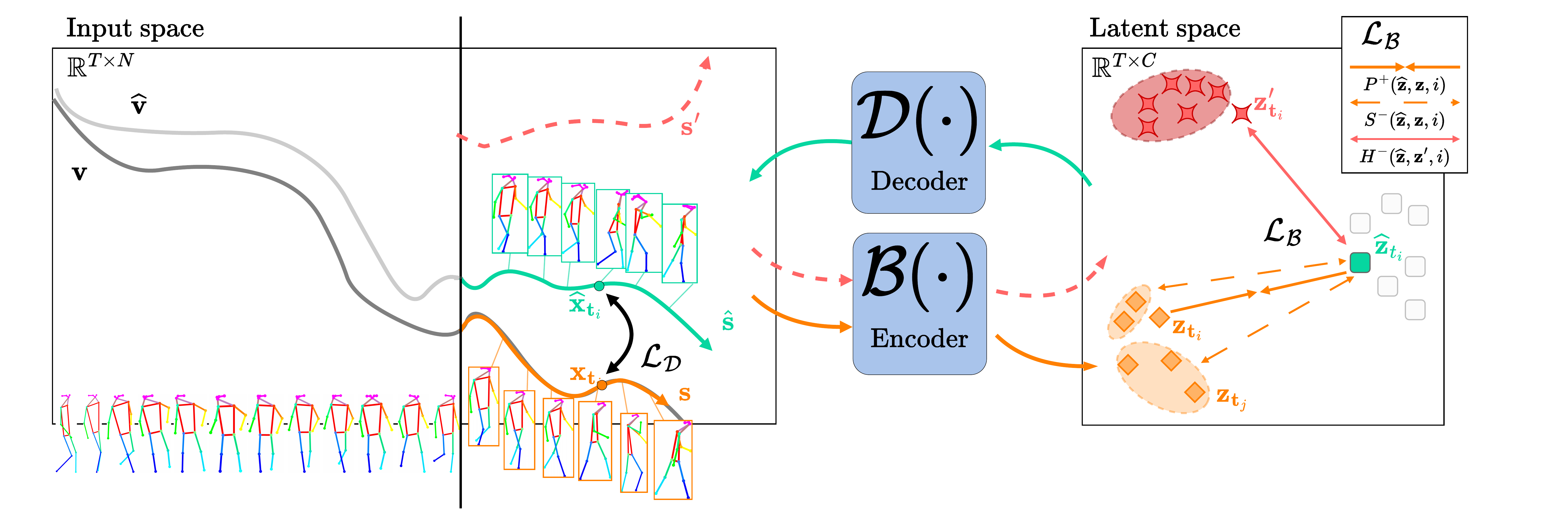}
    \caption{\textbf{Method overview}. Segment $\mathbf{s}$ is occluded from trajectory $\mathbf{v} \in \mathbb{R}^{T \times N}$. Each spatial point $\mathbf{x}$ at time $t_i$ is mapped to $\mathbf{z}_{t_i}$ in the latent space $\mathbb{R}^{T \times C}$ by encoder $\mathcal{B}$. Tensor $\widehat{\mathbf{z}}$ corresponding to expected latent representations is learned contrastively from positive pair $(\widehat{\mathbf{z}}_{t_i},\mathbf{z}_{t_i})$, soft negative pair $(\widehat{\mathbf{z}}_{t_i},\mathbf{z}_{t_j}) \, \text{for } t_j \neq t_i$, and hard negative pair $(\widehat{\mathbf{z}}_{t_i},\mathbf{z}'_{t_i})$. Latent $\mathbf{z}_{t_i}$ of point $\mathbf{x}$ at time $t_i$ is used as a positive pair to $\widehat{\mathbf{z}}_{t_i}$. The remaining projections $\mathbf{z}_{t_j}$ from segment $\mathbf{s}$ are soft negative pairs to $\widehat{\mathbf{z}}_{t_i}$. Latents $\mathbf{z}_{t_i}$ of segments $\mathbf{s}'$ from other trajectories are considered hard negatives. The learned $\widehat{\mathbf{z}}_{t_i}$ is projected back to point $\widehat{\mathbf{x}}$ in the input space $\mathbb{R}^{T \times N}$ through decoder $\mathcal{D}$. Projected points $\widehat{\mathbf{x}}$ in the estimated trajectory $\widehat{\mathbf{v}}$ are compared to their corresponding points $\mathbf{x}$ in $\mathbf{v}$.}
    \label{fig:overview}
    \vspace{-1em}
\end{figure*}

%------------------------------------------------------------------------
\section{Related work}
\label{sec:related}

% General approaches for skeleton action recognition (not anomaly detection)
\noindent
\textbf{Skeleton trajectories modeling}. Motion modeling in videos with skeleton trajectories has been widely studied for tasks such as action recognition~\cite{duan2022revisiting,foo2023unified,plizzari2021skeleton,stergiou2019analyzing}, person re-identification~\cite{elaoud2017analysis,elaoud2020modeling}, human-computer interaction~\cite{li2020learning}, virtual reality~\cite{ciftci2017partially}, and robotics~\cite{bandi2021skeleton}. A large portion of skeleton-based approaches has relied on graph convolutional networks~\cite{cai2021jolo,chen2021channel,chi2022infogcn,gupta2021quo,yan2018spatial} that model sequences as spatio-temporal graphs. Recent approaches have also been based on the representation of trajectories as spatiotemporal volumes through heatmaps~\cite{duan2022revisiting}. Utilizing the continuity of time to create a singular signal for trajectories has also recently gained traction~\cite{suris2022representing,suris2021learning} as it provides a flexible representation of the progression of actions. We adopt a similar approach by minimizing the reconstruction errors between the ground truth trajectory and the predicted trajectory reconstructed by the decoder. Jointly, the correlation of segments at different times and their continuous trajectory is maximized with our model learning to encapsulate trajectory information across different segments.

% prior works on Video Anomaly Detection (Trajectory-based are discussed in detail)
\noindent
\textbf{Video anomaly detection}. The majority of anomaly detection methods in the literature~\cite{feng2021mist,nguyen2019anomaly,sabokrou2017deep,tian2021weakly,yuan2020dlow,wang2021multi,akccay2019skip,hasan2016learning,liu2018future,luo2017remembering,luo2017revisit,ravanbakhsh2017abnormal,zhao2017spatio,flaborea2023contracting,morais2019learning,flaborea2023multimodal,markovitz2020graph,sultani2018real} are autoregressive, where abnormal behavior is only defined by inferring future skeletons from previous ones. However, these approaches are incapable of modeling the continuity of trajectories in full and can only reconstruct a future segment of the trajectory. Some methods use a score function to maximize the embedding space distance between normal and abnormal instances~\cite{feng2021mist,nguyen2019anomaly,sabokrou2017deep,tian2021weakly,yuan2020dlow,wang2021multi}. A different line of approaches learn to reconstruct normal trajectory segments with anomalous segments only present in inference~\cite{akccay2019skip,hasan2016learning,liu2018future,luo2017remembering,luo2017revisit,ravanbakhsh2017abnormal,zhao2017spatio}. Morais~\etal~\cite{morais2019learning} used a recurrent encoder-decoder to predict and reconstruct proceeding frames. Following works have included clustering embeddings of similar classes~\cite{markovitz2020graph} or hyperspherical representations of latents~\cite{flaborea2023contracting}. Luo~\etal~\cite{luo2021normal} represented trajectories as graphs and used Graph Convolutions for predicting future keypoints. Other approaches have jointly studied the tasks of learning the arrow of time, predicting motion irregularity, and predicting object appearance~\cite{georgescu2021anomaly}.  Flaborea~\etal~\cite{flaborea2023multimodal} used a diffusion-based model to synthesize future poses conditioned on past trajectories. Despite strong baseline results, these approaches are not capable of learning to discriminate between trajectories of normal and abnormal behavior beyond the temporal limits of future segments.

% Self-supervision for temporal sequences
\noindent
\textbf{Self supervised representations}. Learning temporal signal representations through contrastive learning has been explored for video~\cite{dave2022tclr,fernando2021anticipating,pan2021videomoco,tong2022videomae}, audio~\cite{ferraro2021enriched,fonseca2021unsupervised,yadav2022learning}, and  pose~\cite{lee2023image,spurr2021self,ziani2022tempclr}. For trajectories, works have focused on using augmentations comparing positive and negative instances~\cite{halawa2022action,li2023self,liu2021social,makansi2021exposing}. A set of recent approaches has also used self-supervision to represent hierarchical relationships between segments~\cite{park2022probabilistic,yang2023self}. Sur{\'\i}s and Vondrick~\cite{suris2022representing} have represented segments through a probability distribution in a latent space. In our work, we use self-supervision to correlate and compare representations of trajectory segments and learned latent representations of normal trajectories.

\section{Method}
\label{sec:method}

In this section, we overview our approach, shown in Figure~\ref{fig:overview}. We first introduce our holistic multitask approach for learning normality in trajectories in Section~\ref{sec:method::holistic}. Each latent space trajectory segment encodes the expected progression of a normal trajectory at different times through positive and negative-pair self-supervision defined in Section~\ref{sec:method::latent_learning}. Latent segment representations are decoded to the input space and compared to normal/abnormal trajectories as detailed in Section~\ref{sec:method::decoder}.

\subsection{Definitions}
\label{sec:method::def}

% Definitions for inputs space, trajectory, segments, and points
We use standard VAD definitions from the literature~\cite{markovitz2020graph,morais2019learning,flaborea2023contracting,flaborea2023multimodal} where trajectories correspond to anomalous events characterized by irregular motions. Trajectories of skeleton keypoints from non-anomalous instances are used to learn normal behaviors, which are then reconstructed and compared to abnormal sequences at inference. We denote the full continuous trajectory over $T$ frames as $\textbf{v} \in \mathbb{R}^{T \times N}$, where $N$ are the spatial coordinates of each point $\mathbf{x}$ over $\mathbf{t} = \{1,...,T\}$ locations. We define a sub-sequence as continuous segment $\mathbf{s}$ over $\widehat{\mathbf{t}}$ temporal locations. 

% Definitions of our framework
The motion information of the entire trajectory $\mathbf{v}$ is encoded from the input space into representation $\mathbf{z}$ within latent space $\mathbb{R}^{T \times C}$. Each point $\mathbf{z}_{t_i}$ in the latent space corresponds to a spatial point of the trajectory at a time $t_i \in \mathbf{t}$. Our goal is to approximate the occluded segment $\mathbf{s}$ and its representations when only part of the trajectory is observed $\mathbf{v} \setminus \mathbf{s} = \{\mathbf{x}: \mathbf{x}\in \mathbf{v} \; \text{and} \; \mathbf{x} \notin \mathbf{s}\}$. We use an encoder $\mathcal{B}$ to encode each point $\mathbf{x}$ at $t_i$ into the latent space $\mathcal{B}(\{\mathbf{x};\mathbf{v} \setminus \mathbf{s}\})$. The latent representations are then combined with a learned tensor to form an estimated latent trajectory. We use a decoder $\mathcal{D}$ to decode each representation at temporal point $t_i$ back to the input space and obtain point $\widehat{\mathbf{x}}$.

\subsection{Multitask Holistic Trajectories}
\label{sec:method::holistic}

% Motivation for holistic representation of trajectories
Typically, a natural choice for the detection of anomalies would be to extrapolate expected motion patterns only for future segments. However, we argue that jointly learning multiple trajectory segments is crucial for both distinguishing anomalies that may occur at different times as well as learning a high-level understanding of the global trajectory. For instance, the future trajectory of a person who is slowing down is less likely to include abnormalities than the past. Additionally, the continuity of trajectories relies on pose tracking with occlusions at both the keypoint and frame levels present at different times. We thus propose a holistic representation of trajectories for past, present, and future segments. Given partial trajectory $\mathbf{v} \setminus \mathbf{s}$ we predict occluded segments $\mathbf{s} = \{\mathbf{x}_{t_i}: t_i \in \widehat{\mathbf{t}}\}$ for three tasks.

% Task future given past
\noindent
\textbf{Predicting the future given the past} (\textcolor{future}{Ftr}). \emph{Future segments} over $\widehat{\mathbf{t}} = \{ T_{Ftr},...,T \}$ are estimated from partial trajectory $\mathbf{v} \setminus \mathbf{s}$ composed of only past segments.

% Task past given future
\noindent
\textbf{Predicting the past given the future} (\textcolor{past}{Pst}). \emph{Past segments} over $\widehat{\mathbf{t}} = \{ 1,...,T_{Pst}\}$ are estimated from partial trajectory $\mathbf{v} \setminus \mathbf{s}$ of only future segments.

% Task present given past and future
\noindent
\textbf{Predicting the present given both past and future} (\textcolor{present}{Prs}). In-between segments over $\widehat{\mathbf{t}} = \{ T_{Pst}+1,...,T_{Ftr}-1 \}$, are referred to as \emph{present segments} and estimated from partial trajectory $\mathbf{v} \setminus \mathbf{s}$ composed of both past and future segments.

We explore these tasks jointly in a multitask training scheme. Latent representations for each of the occluded \textcolor{past}{Pst}, \textcolor{present}{Prs}, and \textcolor{future}{Ftr}, segments are learned jointly and reconstructed back to the input space by the decoder. 

\subsection{Latent Representation Leaning}
\label{sec:method::latent_learning}

% Encoding trajectories to latent space
We use an attention-based encoder $\mathcal{B}$ applied only on the un-occluded part of trajectories $\mathbf{v} \setminus \mathbf{s}$. Similar to~\cite{morais2019learning} we use the trajectories of both skeleton keypoints and the corresponding bounding box corners. We define a learnable tensor $\mathbf{u} = \{ \widehat{\mathbf{z}}_{t_i} : t_i \in \mathbf{t} \}$ of equal size as the full trajectory's representations $|\mathbf{u}| = |\mathcal{B}(\mathbf{v})|$. For each task, we select segment $\mathbf{u}_s = \{ \widehat{\mathbf{z}}_{t_i} : t_i \in \widehat{\mathbf{t}} \}$ corresponding to the occluded $\mathbf{s}$ in $\mathbf{v}$. Tensor segment $\mathbf{u}_s$ is combined with the observed trajectory latents:
\begin{equation}
\label{eq:reorering}
    \mathbf{a} = \Phi(\mathcal{B}(\{\mathbf{x};\mathbf{v} \setminus \mathbf{s}\}) \cup \textbf{u}_{s})
\end{equation}
\noindent
where $\Phi$ is a function that reorders $\mathcal{B}(\{\mathbf{x};\mathbf{v} \setminus \mathbf{s}\}) \cup \textbf{u}_{s}$ given the temporal location of $\mathbf{s}$ within $\mathbf{v}$. We run inference to also obtain the representations of the entire sequence $\mathbf{z} = \mathcal{B}(\mathbf{v})$. 

% Learning of latent (positive/negative pairs)
Consequently, we want each of the learned vectors $\widehat{\mathbf{z}}_{t_i} \in \mathbf{u}_s$ to be drawn closer to the corresponding trajectory representations $\mathbf{z}_{t_i} \in \mathcal{B}(\mathbf{v})$ at temporal point $t_i$ while being pushed away from representations at other trajectory points or other random points. To encourage this, we define an objective based on latent space positive and negative pairs.

% Hard positives
\noindent
\textbf{Positive pairs}. We wish to minimize the distance between learned vector $\widehat{\mathbf{z}}_{t_i}$ and the trajectory representation $\mathbf{z}_{t_i}$ at $t_i$. We thus define pair $P^{+}(\widehat{\mathbf{z}},\mathbf{z},i)= \|\widehat{\mathbf{z}}_{t_i} - \mathbf{z}_{t_i} \|_2$ as a \emph{positive}.

% Soft negatives
\noindent
\textbf{Soft negative pairs}. Given the remaining representations $\mathbf{z}_{t_j}$ at temporal locations $t_j \neq t_i$ we want to maximize their distance to $\widehat{\mathbf{z}}_{t_i}$. However, as $\mathbf{z}_{t_j}$ belongs to the same trajectory for which $\widehat{\mathbf{z}}_{t_i}$ is an estimate of, they are bound to include some similarities. We thus treat $(\widehat{\mathbf{z}}_{t_i},\mathbf{z}_{t_j})$ as \emph{soft negative} pairs for which their difference is regularized by their temporal distance. We want to penalize less pairs of points that are temporally close comparatively to pairs of points further away: $\mathcal{R}(t_i,t_{i+1}) < \mathcal{R}(t_i,t_{i+2})$. The soft negative pairs penalty is defined as:
\begin{gather}
\begin{aligned}
\label{eq:soft_negatives}
   S^-(\widehat{\mathbf{z}},\mathbf{z},i) &= \sum_{t_{j} \in \widehat{\mathbf{t}}} \mathcal{R}(t_i,t_{j}) \,  \|\widehat{\mathbf{z}}_{t_i} - \mathbf{z}_{t_j}\|_2, \; \text{where} \\
    \mathcal{R}(t_i,t_{j}) &= \frac{\beta \, \|t_i-t_{j}\|_1}{\text{max}(\|t_i-t_{k}\|_1 : t_k \in \widehat{\mathbf{t}})}
\end{aligned}
\end{gather}
\noindent
where $\beta$ is a hyperparameter used to adjust the penalization.

% Hard negatives
\noindent
\textbf{Hard negative pairs}. We select latents of segments $\mathbf{s}'$ from other trajectories as \emph{hard negative} pairs. As each segment $\mathbf{s}$ is unique to trajectory $\mathbf{v}$, no representation $\mathbf{z}'$ from segment $\mathbf{s}'$ should correspond to $\widehat{\mathbf{z}}_{t_i}$. Thus, we also maximize their distance $H^-(\widehat{\mathbf{z}},\mathbf{z}',i)=\|\widehat{\mathbf{z}}_{t_i} - \mathbf{z}'_{t_i}\|_2$.

% Encoder loss
We use a self-supervised triplet-loss $\mathcal{L}_{\mathcal{B}}$ with an additional \emph{soft negative} pair term. This forces encoder $\mathcal{B}$ to attain estimated representations $\widehat{\mathbf{z}}_{t_i}$ close to latents $\mathbf{z}_{t_i}$ at time $t_i$. It also reduces the resemblance to remaining representations $\mathbf{z}_{t_j}$ from other temporal locations as well as representations $\mathbf{z}'_{t_i}$ from segments $\mathbf{s}'$ of other trajectories.
\begin{equation}
\begin{split}
\label{eq:loss_encoder}
    \mathcal{L}_{\mathcal{B}} = \sum_{ t_i \in \widehat{\mathbf{t}}} \text{max}(P^{+}(\widehat{\mathbf{z}},\mathbf{z},i) - S^-(\widehat{\mathbf{z}},\mathbf{z},i) \\ - H^-(\widehat{\mathbf{z}},\mathbf{z}',i) + \gamma , 0)
\end{split}
\end{equation}
\noindent
where $\gamma$ is a margin hyperparameter.

\addtocounter{footnote}{0}
\begin{table*}[t]
\centering
\caption{\textbf{AUC performance of VAD methods on HR-STC, HR-Avenue, HR-UBnormal, and UBnormal}. \textcolor{future}{Ftr} is the common task addressed by all VAD methods in the literature. Top performances for each task and for each dataset are in \textbf{bold}.}
\resizebox{.98\linewidth}{!}{%
\begin{tabular}{l c c c c c c c c c c c c c c c}
\hline
\multirow{3}{*}{}&
\multicolumn{3}{c}{\textbf{HR-STC}} &&
\multicolumn{3}{c}{\textbf{HR-Avenue}} &&
\multicolumn{3}{c}{\textbf{HR-UBnormal}} &&
\multicolumn{3}{c}{\textbf{UBnormal}}
\tstrut \bstrut \\
\cline{2-4} \cline{6-8} \cline{10-12} \cline{14-16}
\tstrut & \textcolor{future}{\textbf{Ftr}} & \textcolor{present}{\textbf{Prs}} & \textcolor{past}{\textbf{Pst}} && \textcolor{future}{\textbf{Ftr}} & \textcolor{present}{\textbf{Prs}} & \textcolor{past}{\textbf{Pst}} && \textcolor{future}{\textbf{Ftr}} & \textcolor{present}{\textbf{Prs}} & \textcolor{past}{\textbf{Pst}}&& \textcolor{future}{\textbf{Ftr}} & \textcolor{present}{\textbf{Prs}} & \textcolor{past}{\textbf{Pst}} \bstrut \\
\hline
GEPC~\cite{markovitz2020graph} & 
74.8 & - & - && 
58.1 & - & - &&
55.2 & - & - &&
53.4 & - & - \tstrut \\
MPED-RNN~\cite{morais2019learning} & 
75.4 & 69.8\footnote{note} &72.1\footref{note} && 
86.2 & 81.4\footref{note} & 83.8\footref{note} &&
61.2 & 59.3\footref{note} & 61.1\footref{note} &&
60.6 & 58.5\footref{note} & 60.1\footref{note} \\
COSKAD~\cite{flaborea2023contracting} & 
77.1 & - & - && 
87.3 & - & - &&
65.5 & - & - &&
65.0 & - & -\\
MoCoDAD~\cite{flaborea2023multimodal} & 
77.6 & - & - &&
89.0 & - & - &&
\textbf{68.4} & - & - &&
\textbf{68.3} & - & -\\
\textbf{TrajREC (ours)} & 
\textbf{77.9} & \textbf{73.5} & \textbf{75.7} && 
\textbf{89.4} & \textbf{86.3} & \textbf{87.6} &&
68.2 & \textbf{64.1} & \textbf{66.8} &&
68.0 & \textbf{63.6} & \textbf{66.4} \\
\end{tabular}
}
\label{tab:sota}
\vspace{-1.1em}
\end{table*}

\subsection{Trajectory segment reconstruction}
\label{sec:method::decoder}

We wish to decode the representations of the trajectory's spatial points from latent space $\mathbb{R}^{T \times C}$ back into the input space $\mathbb{R}^{T \times N}$. Decoder $\mathcal{D}$ takes $\mathbf{a}$ from~(\ref{eq:reorering}) and projects representations $\mathcal{B}(\{\mathbf{x};\mathbf{v} \setminus \mathbf{s}\})$ of the un-occluded trajectory $\mathbf{v} \setminus \mathbf{s}$ alongside the occluded learned segment $\mathbf{u}_s$, back to $\mathbb{R}^{T \times N}$. As both ground truth trajectory $\mathbf{v}$ and estimated $\widehat{\textbf{v}} = \mathcal{D}(\textbf{a})$ are available, the decoder is explicitly trained on reconstruction and not extra/interpolation. 
\begin{equation}
\begin{split}
\label{eq:loss_decoder}
    \mathcal{L}_{\mathcal{D}} = \frac{1}{T}\sum_{t_i \in \mathbf{t}} \|\widehat{\mathbf{x}}_{t_i} - \mathbf{x}_{t_i}\|_2
\end{split}
\end{equation}

\noindent
where $\widehat{\mathbf{x}} \in \widehat{\mathbf{v}}$ and $\mathbf{x} \in \mathbf{v}$. The decoder is trained to regress the error between reconstructed $\widehat{\mathbf{x}}$ and ground truth $\mathbf{x}$ at temporal location $t_i$ in the input space $\mathbb{R}^{T \times N}$. At inference, only decoded points $\widehat{\mathbf{x}} \in \widehat{\mathbf{s}}$ are compared to points $\mathbf{x} \in \mathbf{s}$.

%Combined loss
We combine the encoder and decoder losses from~(\ref{eq:loss_encoder}) and~(\ref{eq:loss_decoder}), and define our multitask learning objective:
\begin{equation}
\label{eq:multitask}
    \mathcal{L} = \mathcal{L}_{\mathcal{B}}+\lambda\mathcal{L}_{\mathcal{D}}
\end{equation}
\noindent
where $\lambda$ is a hyperparameter that is tuned according to the trajectory belonging to either a skeleton keypoint or a bounding box corner. At each training step, we extrapolate/interpolate an equal number of trajectory segments for each \textcolor{past}{Pst}, \textcolor{present}{Prs}, and \textcolor{future}{Ftr} task before a gradient update. As each task uses a different segment $\mathbf{u}_s$, updates are computed accumulatively for the entirety of $\mathbf{u}$ to stabilize training.

%------------------------------------------------------------------------
\section{Experiments}
\label{sec:experiments}

The datasets used, alongside implementation and training details are described in Section~\ref{sec:experiments::details}. We compare our proposed approach to state-of-the-art models in Section~\ref{sec:experiments::results}. We show qualitative results on all three tasks in Section~\ref{sec:experiments::qualitative} followed by ablation studies in Section~\ref{sec:experiments::ablation}. 

\subsection{Experimental details}
\label{sec:experiments::details}

% Datasets overview
\noindent
\textbf{Datasets}. We report our framework's performance over a diverse set of VAD datasets. Human-Related ShanghaiTech Campus (\emph{HR-STC})~\cite{luo2017revisit,morais2019learning} consists of 13 scenes with 303 train and 101 test videos containing 130 anomalous events. \emph{HR-Avenue}~\cite{lu2013abnormal,morais2019learning} includes a single scene with 16 train and 21 test videos. \emph{UBnormal}~\cite{acsintoae2022ubnormal} is a collection of 29 synthetic scenes synthesized with Cinema4D and natural backgrounds with 186 normal train and 211 test videos. \emph{HR-UBnormal}~\cite{acsintoae2022ubnormal,flaborea2023contracting} is a subset of UBnormal introduced by \cite{flaborea2023contracting} that includes only HR anomalies in the test set. For all datasets, we use the provided skeleton keypoints and bounding box coordinates detected with AlphaPose~\cite{fang2022alphapose}. Following works in the literature~\cite{acsintoae2022ubnormal,flaborea2023contracting,flaborea2023multimodal,markovitz2020graph,morais2019learning,liu2018future,lu2013abnormal} we use the Area Under the Receiver Operating Characteristic Curve (AUC) score as our performance metric.

% training/method settings
\noindent
\textbf{Model settings}. We train our model jointly on all three tasks; \textcolor{future}{Ftr}, \textcolor{present}{Pst}, and \textcolor{past}{Pst}. We directly compare to the state-of-the-art for \textcolor{future}{Ftr} as it has been primarily used for VAD, and provide baselines for \textcolor{present}{Pst}, and \textcolor{past}{Pst}. 
Overall, we employ two STTFormers~\cite{qiu2022spatio}, one as the encoder $\mathcal{B}$ for projecting trajectory points to $\mathbb{R}^{T \times C}$ and the second is inversed and used as the decoder to project points back to $\mathbb{R}^{T \times N}$. Unless otherwise specified, we use $T=18$ trajectory frames with segment lengths of $|\widehat{\mathbf{t}}|=6$. This corresponds to input sequences of $18 \times (17+4) \times 2$, where 17 is the number of skeleton joints, 4 is the corners of the bounding box of the skeleton, and 2 corresponds to their spatial locations. The size of the learned latent tensor $\mathbf{u}$ is $|\mathbf{u}| = 18 \times 256$. We train with $1e^{-4}$ base learning rate and a batch size of 512. We use $\beta=0.001$, and $\gamma=0.1$, with $\lambda=5$ for joint and $\lambda=3$ for bounding box corner trajectories. An overview of AUC scores with different hyperparameter combinations is shown in \textsection \textcolor{red}{S1} in the supplementary material.   

\footnotetext{\label{note}Inhouse retrained for the specific task.}

\subsection{Comparative results}
\label{sec:experiments::results}

% Comparisons and baseline definition
In Table~\ref{tab:sota} we report AUC scores on the three tasks; extrapolation of future keypoints from past keypoints \textcolor{future}{Ftr}, extrapolation of past keypoints from future keypoints \textcolor{past}{Pst}, and interpolation of in-between/present keypoints from both future and past keypoints \textcolor{present}{Prs}. Current models are not capable of jointly extra/interpolating keypoints and are instead only limited to future extrapolation \textcolor{future}{Ftr}. We select MPED-RNN~\cite{morais2019learning} as a baseline for the previously unexplored past extrapolation \textcolor{past}{Pst} and present interpolation \textcolor{present}{Prs}, due to its popular use as a VAD baseline and open-source implementation. The design of MPED-RNN does not allow multitask learning so we retrain individually for \textcolor{past}{Pst} and \textcolor{present}{Prs}. 

\begin{figure*}[t]
    \centering
    \begin{subfigure}[b]{\textwidth}
        \includegraphics[width=\linewidth]{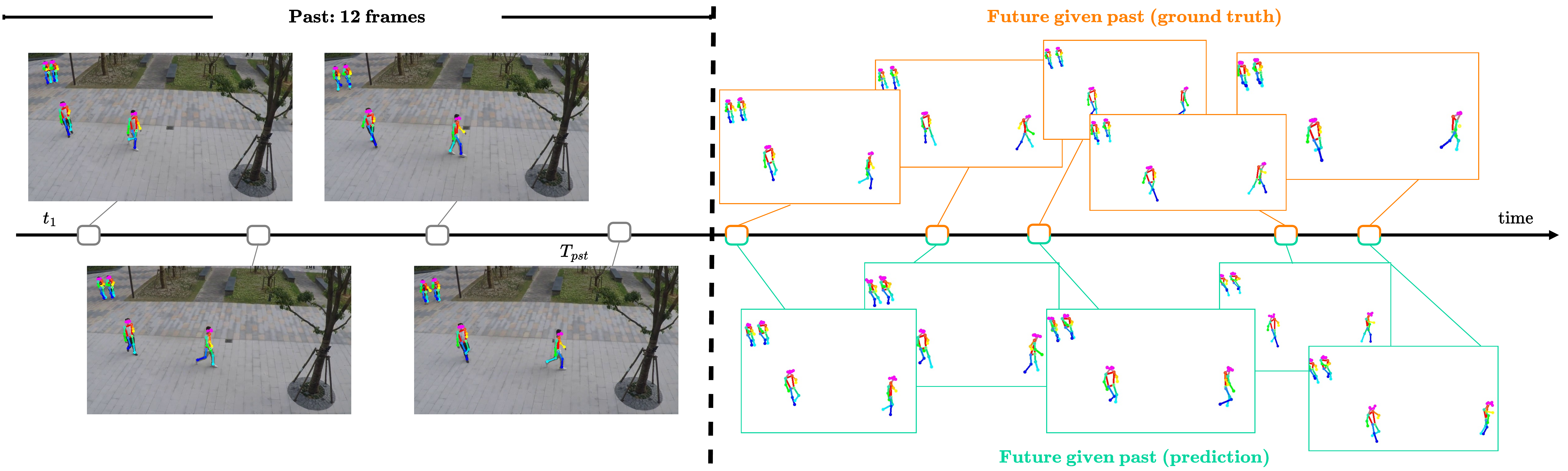}
        \caption{\textbf{Predicting the future given the past} (\textcolor{future}{Ftr}) \textbf{for a normal trajectory}.}
    \label{fig:examples:extrapolation_future_normal}
    \end{subfigure}
    \begin{subfigure}[b]{\textwidth}
        \vspace{.4em}
        \includegraphics[width=\linewidth]{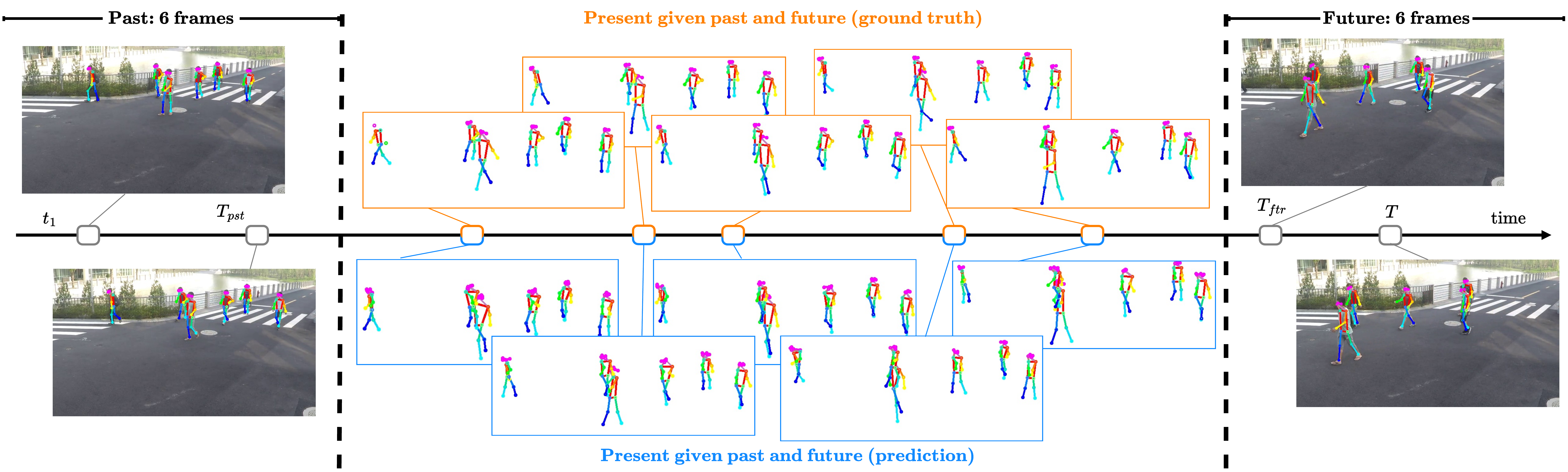}
        \caption{\textbf{Predicting the present given both the past and future} (\textcolor{present}{Prs}) \textbf{for a normal trajectory}.}
    \label{fig:examples:interpolation_normal}
    \end{subfigure}
    \begin{subfigure}[b]{\textwidth}
        \vspace{.4em}
        \includegraphics[width=\linewidth]{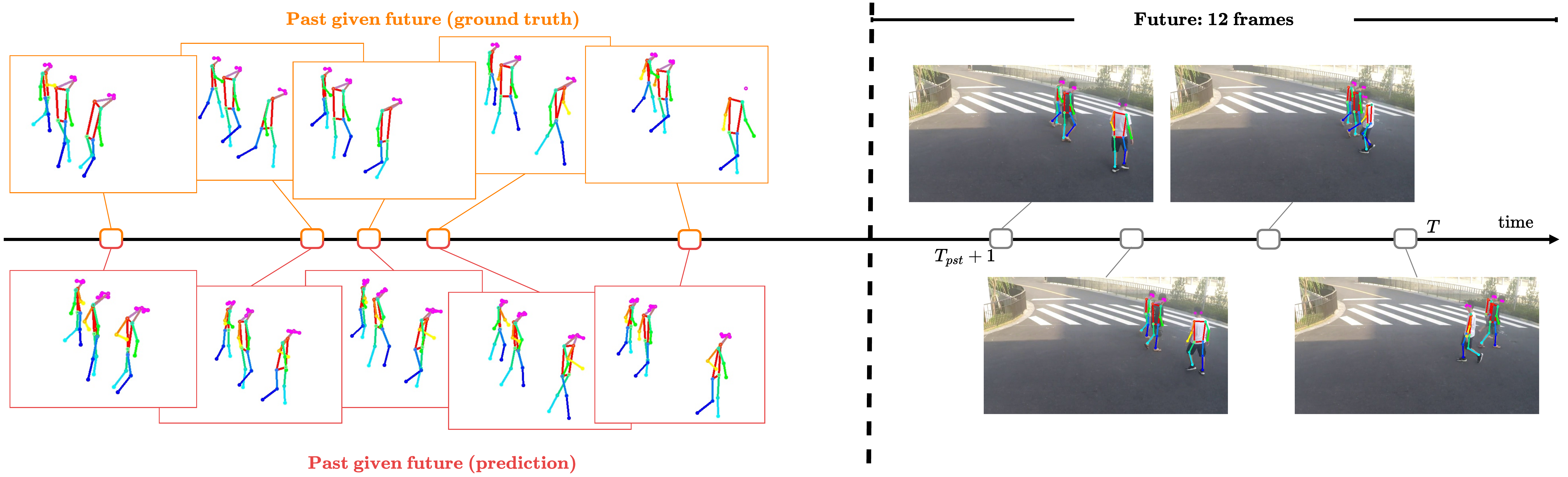}
        \caption{\textbf{Predicting the past given the future} (\textcolor{past}{Pst}) \textbf{for a normal trajectory}.}
    \label{fig:examples:extrapolation_past_normal}
    \end{subfigure}
    \caption{\textbf{Normal and abnormal skeleton trajectory reconstruction for HR-STC}. Input trajectories are keypoints from 18 frames with each of the predicted segments for \textcolor{future}{Ftr}, \textcolor{present}{Prs}, and \textcolor{past}{Pst} task being 6 frames. Both extrapolations of future and past segments as well as the interpolation of in-between segments are sensible predictions of normal behaviors across all skeletons in each scene.}
    \label{fig:examples}
    \vspace{-1em}
\end{figure*}
\begin{figure*}[t]\ContinuedFloat
    \centering
    \begin{subfigure}[b]{\textwidth}
        \includegraphics[width=\linewidth]{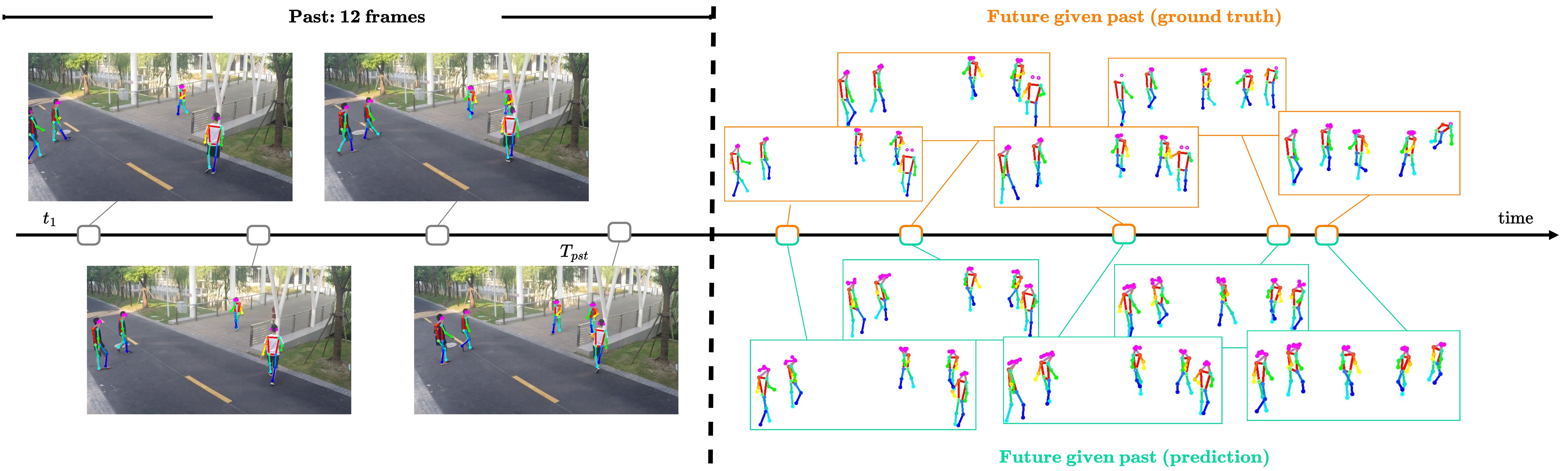}
        \caption{\textbf{Predicting the future given the past} (\textcolor{future}{Ftr}) \textbf{for an abnormal trajectory}.}
    \label{fig:examples:extrapolation_future_abnormal}
    \end{subfigure}
    \begin{subfigure}[b]{\textwidth}
        \vspace{.4em}
        \includegraphics[width=\linewidth]{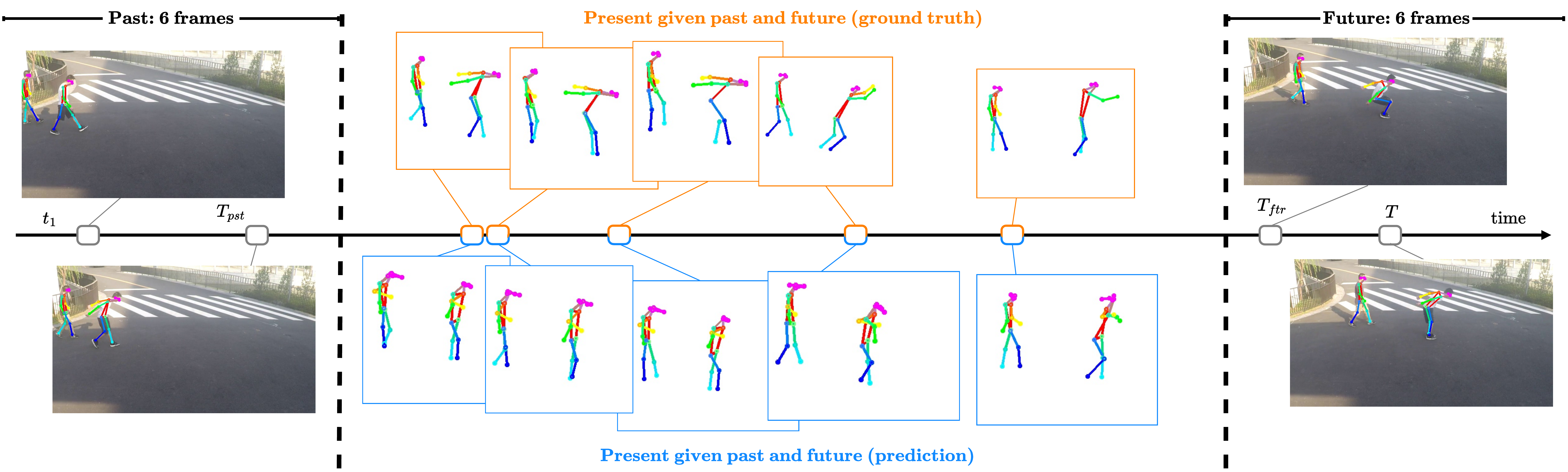}
        \caption{\textbf{Predicting the present given both the past and future} (\textcolor{present}{Prs}) \textbf{for an abnormal trajectory}.}
    \label{fig:examples:interpolation_abnormal}
    \end{subfigure}
    \begin{subfigure}[b]{\textwidth}
        \vspace{.4em}
        \includegraphics[width=\linewidth]{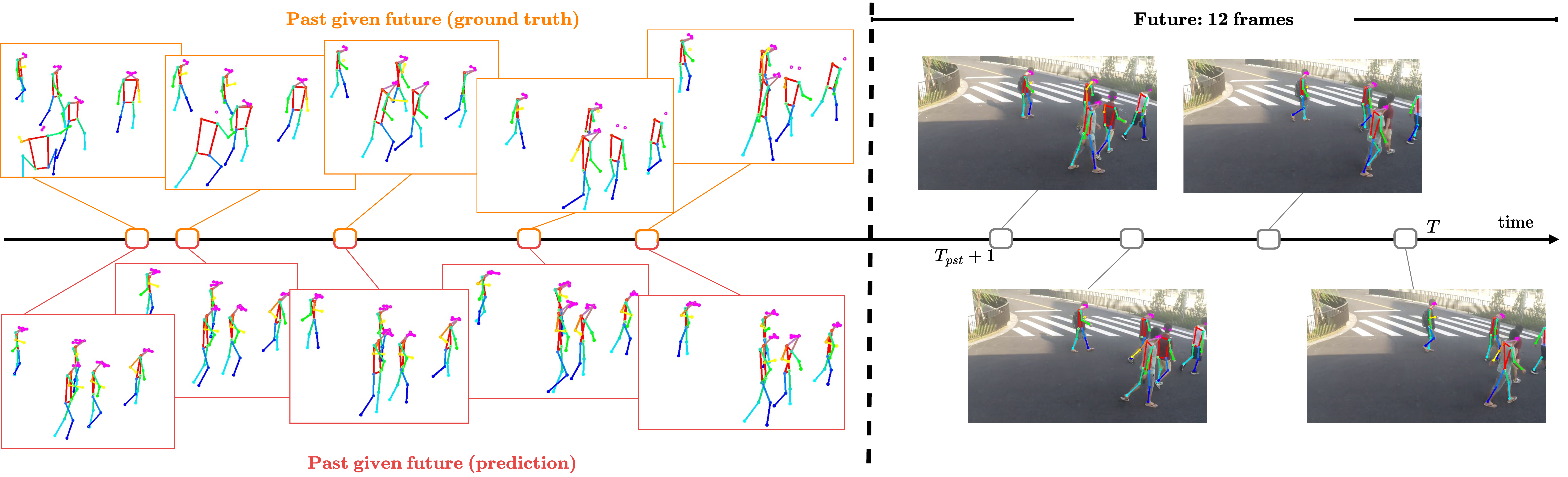}
        \caption{\textbf{Predicting the past given the future} (\textcolor{past}{Pst}) \textbf{for an abnormal trajectory}.}
    \label{fig:examples:extrapolation_past_abnormal}
    \end{subfigure}
    \caption{\textbf{Normal and abnormal skeleton trajectory reconstruction for HR-STC}. Input trajectories are keypoints from 18 frames with each of the predicted segments for \textcolor{future}{Ftr}, 
\textcolor{present}{Prs}, and \textcolor{past}{Pst} task being 6 frames. Both extrapolations of future and past segments as well as the interpolation of in-between segments are sensible predictions of normal behaviors across all skeletons in each scene (continued).}
\vspace{-1em}
\end{figure*}

\noindent
\textbf{HR-STC}. Across the \textcolor{future}{Ftr} task we show that our proposed approach outperforms all other single-task approaches. We achieve a +2.5 AUC score improvement over the baseline~\cite{morais2019learning} and +0.3 over the best-performing single-task model~\cite{flaborea2023multimodal}. For \textcolor{past}{Pst} our model achieves a 75.7 AUC score outperforming~\cite{morais2019learning} by +3.6. Similar score increases are also observed for the \textcolor{present}{Prs} task with a +3.7 improvement over the baseline. Our method's flexibility enables learning and inferring holistic representations of the entire trajectory which can evidently benefit all three tasks jointly.

\noindent
\textbf{HR-Avenue}. Compared to the baseline, we observe a +3.2, +3.8, and +4.9 improvement in the AUC score for the \textcolor{future}{Ftr}, \textcolor{past}{Pst}, and \textcolor{present}{Prs} tasks respectively. Our method also surpasses the previous state-of-the-art models~\cite{flaborea2023multimodal} on the \textcolor{future}{Ftr} task with a +0.4 increase in the AUC score. 

\noindent
\textbf{HR-UBnormal}. We observe that the relative increase in performance also varies across datasets with +7.0 for the \textcolor{future}{Ftr}, +4.8 for \textcolor{present}{Prs}, and +5.7 for \textcolor{past}{Pst} AUC score increase over the baseline. We believe this to be due to the scene diversity of the dataset. Due to UBnormal using synthetic data, we speculate that methods based on generative models such as~\cite{flaborea2023multimodal} can better model motions in synthetic trajectories.

\noindent
\textbf{UBnormal}. The results on the full UBnormal dataset follow the same trends as those from the HR-UBnormal subset. Relative to the baseline, the largest improvement across tasks is observed for \textcolor{future}{Ftr} with a +7.4 AUC score increase followed by \textcolor{past}{Pst} with +6.3 improvement. For the \textcolor{present}{Prs} task we note a +5.1 improvement in the AUC score.

\begin{table}[t]
\centering
\caption{\textbf{AUC scores on HR-STC and HR-Avenue over different trajectory lengths}. In all settings, the predicted segments are equal to a third of the trajectory's length. }
\resizebox{\linewidth}{!}{%
\begin{tabular}{l c c c c c c c }
\hline
\multirow{3}{*}{}&
\multicolumn{3}{c}{\textbf{HR-STC}} &&
\multicolumn{3}{c}{\textbf{HR-Avenue}} 
\tstrut \bstrut \\
\cline{2-4} \cline{6-8}
\tstrut & \textcolor{future}{\textbf{Ftr}} & \textcolor{present}{\textbf{Prs}} & \textcolor{past}{\textbf{Pst}} && \textcolor{future}{\textbf{Ftr}} & \textcolor{present}{\textbf{Prs}} & \textcolor{past}{\textbf{Pst}} \bstrut \\
\hline
\rowcolor{LightGrey} \multicolumn{8}{l}{Trajectory length $T=18, \; \text{Segment length } |\widehat{\mathbf{t}}|=6$} \tstrut \bstrut \\
Baseline~\cite{morais2019learning} & 
75.4 & 69.8 & 72.1 && 
86.2 & 81.4 & 83.8 \tstrut \bstrut \\
\textbf{TrajREC (ours)} & 
\textbf{77.9} & \textbf{73.5} & \textbf{75.7} && 
\textbf{89.4} & \textbf{86.3} & \textbf{87.6} \tstrut \bstrut \\
\hline
\rowcolor{LightGrey} \multicolumn{8}{l}{Trajectory length $T=24, \; \text{Segment length } |\widehat{\mathbf{t}}|=8$} \tstrut \bstrut \\
Baseline~\cite{morais2019learning} & 
72.9 & 67.2 & 69.8 && 
84.0 & 80.8 & 82.6 \tstrut \bstrut \\
\textbf{TrajREC (ours)} & 
\textbf{76.4} & \textbf{72.7} & \textbf{74.7} && 
\textbf{88.5} & \textbf{85.7} & \textbf{86.3}  \tstrut \bstrut \\
\hline
\rowcolor{LightGrey} \multicolumn{8}{l}{Trajectory length $T=36, \; \text{Segment length } |\widehat{\mathbf{t}}|=12$} \tstrut \bstrut \\
Baseline~\cite{morais2019learning} & 
69.2 & 66.3 & 67.4 && 
81.1 & 78.6 & 80.1 \tstrut \bstrut \\
\textbf{TrajREC (ours)} & 
\textbf{75.3} & \textbf{72.5} & \textbf{73.3} && 
\textbf{86.8} & \textbf{84.1} & \textbf{84.7}  \tstrut \bstrut \\
\end{tabular}
}
\label{tab:seq_len}
\vspace{-1em}
\end{table}

\begin{table}[t]
\centering
\caption{\textbf{Ablations on HR-STC} with different learning scheme settings across different occluded segment $\widehat{\mathbf{t}}$ sizes. Larger $|\widehat{\mathbf{t}}|$ corresponds to smaller visible trajectories and reconstructed segments of larger lengths. The temporal length in all settings is $|T|=18$.}
\resizebox{.9\linewidth}{!}{%
\begin{tabular}{l l c c c }
\hline
 $|\widehat{\mathbf{t}}|$ & Method & \textcolor{future}{\textbf{Ftr}} & \textcolor{present}{\textbf{Prs}} & \textcolor{past}{\textbf{Pst}}  \tstrut \bstrut \\
\hline
\multirow{5}{*}{3} & Baseline~\cite{morais2019learning}  & 76.1 & 70.4 & 72.2 \tstrut \\
                     & \textbf{TrajREC} w/o $H^-$ \textbf{(ours)}  & 77.2 & 71.9 & 74.7 \\
                     & \textbf{TrajREC} w/o $S^-$ \textbf{(ours)}  & 77.5  & 73.0 & 75.6 \\
                     & \textbf{TrajREC single task (ours)} & 77.8 & 73.5 & 75.4 \\
                     & \textbf{TrajREC (ours)} & \textbf{78.1} & \textbf{73.7} & \textbf{76.0} 
 \bstrut\\
\hline
\multirow{5}{*}{6} & Baseline~\cite{morais2019learning}  & 75.4 & 69.8 & 72.1 \tstrut \\
                     & \textbf{TrajREC w/o} $H^-$ \textbf{(ours)}  & 76.9 & 71.4 & 74.0 \\
                     & \textbf{TrajREC w/o} $S^-$ \textbf{(ours)}  & 77.2 & 71.8 & 75.2 \\
                     & \textbf{TrajREC single task (ours)}  & 77.4 & 73.1 & 75.6 \\
                     & \textbf{TrajREC (ours)}  & \textbf{77.9} & \textbf{73.5} & \textbf{75.7} \bstrut \\
\hline
\multirow{5}{*}{9} & Baseline~\cite{morais2019learning}  & 72.9 & 69.1 & 71.7 \tstrut \\
                     & \textbf{TrajREC w/o} $H^-$ \textbf{(ours)}  & 75.2 & 70.9 & 73.8 \\
                     & \textbf{TrajREC w/o} $S^-$ \textbf{(ours)}  & 75.6 & 71.8 & 74.5 \\
                     & \textbf{TrajREC single task (ours)}  & 75.4 & 71.4 & 74.9 \\
                     & \textbf{TrajREC (ours)}  & \textbf{76.2} & \textbf{72.8} & \textbf{76.0} \bstrut \\
\hline
\multirow{5}{*}{12} & Baseline~\cite{morais2019learning}  & 70.6 & 66.8 & 69.4 \tstrut \\
                     & \textbf{TrajREC w/o} $H^-$ \textbf{(ours)}  &  74.1 & 70.4 & 73.7 \\
                     & \textbf{TrajREC w/o} $S^-$ \textbf{(ours)}  & 74.7 & 71.3 & 74.2 \\
                     & \textbf{TrajREC single task (ours)}  & 72.9 & 69.9 & 72.5 \\
                     & \textbf{TrajREC (ours)}  & \textbf{75.9} & \textbf{72.7} & \textbf{75.9} \bstrut \\
\end{tabular}
}
\label{tab:pairs}
\vspace{-1.1em}
\end{table}

\subsection{Qualitative results}
\label{sec:experiments::qualitative}

% Qualitative results based on Figures 3 and 4
In Figure~\ref{fig:examples} we demonstrate examples of reconstructed trajectory segments for normal and anomalous sequences. Extrapolated and interpolated segments of normal trajectories for \textcolor{future}{Ftr}, \textcolor{present}{Prs}, and \textcolor{past}{Pst} are shown in Figures~
\ref{fig:examples:extrapolation_future_normal},~\ref{fig:examples:interpolation_normal}, and~\ref{fig:examples:extrapolation_past_normal}. For all tasks, the model is capable of learning continual sequences with the motions in reconstructed skeletons progressing smoothly. In instances where not all joints are visible or recognized in the input (Figure~\ref{fig:examples:extrapolation_past_normal}) the spatial locations of the predicted joints are reasonable in relation to the neighboring observed keypoints. In Figures~
\ref{fig:examples:extrapolation_future_abnormal},~\ref{fig:examples:interpolation_abnormal}, and~\ref{fig:examples:extrapolation_past_abnormal} we show reconstructions for segments that include anomalies. As the behaviors and motions of these segments are not part of the training set, extrapolated or interpolated segments can be easily detected as anomalous based on the high reconstruction error. In extrapolation tasks, the reconstructed trajectory segments resemble motions of expected normal behavior. In contrast, interpolation requires infilling of occluded segments based on the given past and future keypoints. As proceeding and succeeding frames may also be anomalous (Figure~\ref{fig:examples:interpolation_abnormal}), the reconstructions produced are of low confidence, showing that no normal segment can exist when the frames at the start and end are anomalous. Additional examples are available in \textsection \textcolor{red}{S2}.

\subsection{Ablation studies}
\label{sec:experiments::ablation}

In this section, we ablate over trajectory lengths reporting model performance on segments of increased durations. Additionally, we evaluate our method over different learning schemes and single-task settings. Finally, we present AUC scores over multiple runs with different initializations and show reconstruction examples for each run.

\noindent
\textbf{Trajectory and segment length}. Trajectories and their corresponding segments can vary in length. We evaluate our method on HR-STC and HR-Avenue over multiple trajectory lengths $|T| \in \{ 18, 24, 36 \}$ while keeping the size of the occluded segment equal to a third of the trajectory. As shown in Table~\ref{tab:seq_len}, our model consistently outperforms the baseline~\cite{morais2019learning} across tasks and trajectory lengths. An average -2.0 and -2.6 drop in the AUC scores across tasks is observed for HR-STC and HR-Avenue respectively when increasing $|T| = 18 \rightarrow |T| = 36$. This drop is more prevalent for the baseline with -4.8 and -3.9 for HR-STC and HR-Avenue. Thus, the holistic representation and multitask learning of trajectories is a better strategy for the reconstruction of segments from trajectories with longer durations.

% Moderate reductions in AUC score based on larger occluded segments
\noindent
\textbf{Occlusion length}. We compare the reconstruction performance of our approach at different occlusion lengths in Table~\ref{tab:pairs}. We maintain the temporal resolution of the input $|T|=18$ and occlude segments of progressively increasing lengths; $|\widehat{\mathbf{t}}|=\{3,6,9,12\}$. Larger occlusions correspond to less of the input being observed. Models trained with multitask learning demonstrate only moderate decreases in their AUC score as the length of the occluded segments increases. We note that single-task models including both the baseline and our model trained individually per task $\textbf{u}_s = \textbf{u}$, show a notable performance decrease when the occluded segment's length is increased, as they do not learn a high-level understanding of the global trajectory.

% Learning ablations
\noindent
\textbf{Soft-hard negative pairs}. We additionally include ablations on the effect of hard negative $H^-$ and soft negative $S^-$ pairs to the AUC score in Table~\ref{tab:pairs}. As shown, the removal of either $H^-$ and $S^-$ results in an overall decrease in AUC scores across tasks and occluded segment sizes. Marginally larger AUC score reductions are observed when $H^-$ is removed which we believe is due to points from different trajectories being much stronger negative pairs than points from the current trajectory.

\begin{table}[t]
\centering
\caption{\textbf{AUC scores on HR-STC and HR-Avenue over multiple runs}. The best run is denoted with (best).}
\resizebox{\linewidth}{!}{%
\begin{tabular}{l c c c c c c c }
\hline
\multirow{3}{*}{}&
\multicolumn{3}{c}{\textbf{HR-STC}} &&
\multicolumn{3}{c}{\textbf{HR-Avenue}} 
\tstrut \bstrut \\
\cline{2-4} \cline{6-8}
\tstrut & \textcolor{future}{\textbf{Ftr}} & \textcolor{present}{\textbf{Prs}} & \textcolor{past}{\textbf{Pst}} && \textcolor{future}{\textbf{Ftr}} & \textcolor{present}{\textbf{Prs}} & \textcolor{past}{\textbf{Pst}} \\
\hline
\textbf{run1 (best)} & 
\textbf{77.9} & \textbf{73.5} & \textbf{75.7} && 
\textbf{89.4} & \textbf{86.3} & \textbf{87.6} \tstrut \bstrut \\
\textbf{run2} & 
77.6 & 73.4 & 75.2 && 
89.4 & 86.0 & 87.1  \tstrut \bstrut \\
\textbf{run3} & 
77.9 & 73.2 & 75.4 && 
88.8 & 86.1 & 87.3 \tstrut \bstrut \\
\end{tabular}
}
\label{tab:multi_run}
\vspace{-1.1em}
\end{table}

\begin{figure}
    \centering
    \includegraphics[width=\linewidth]{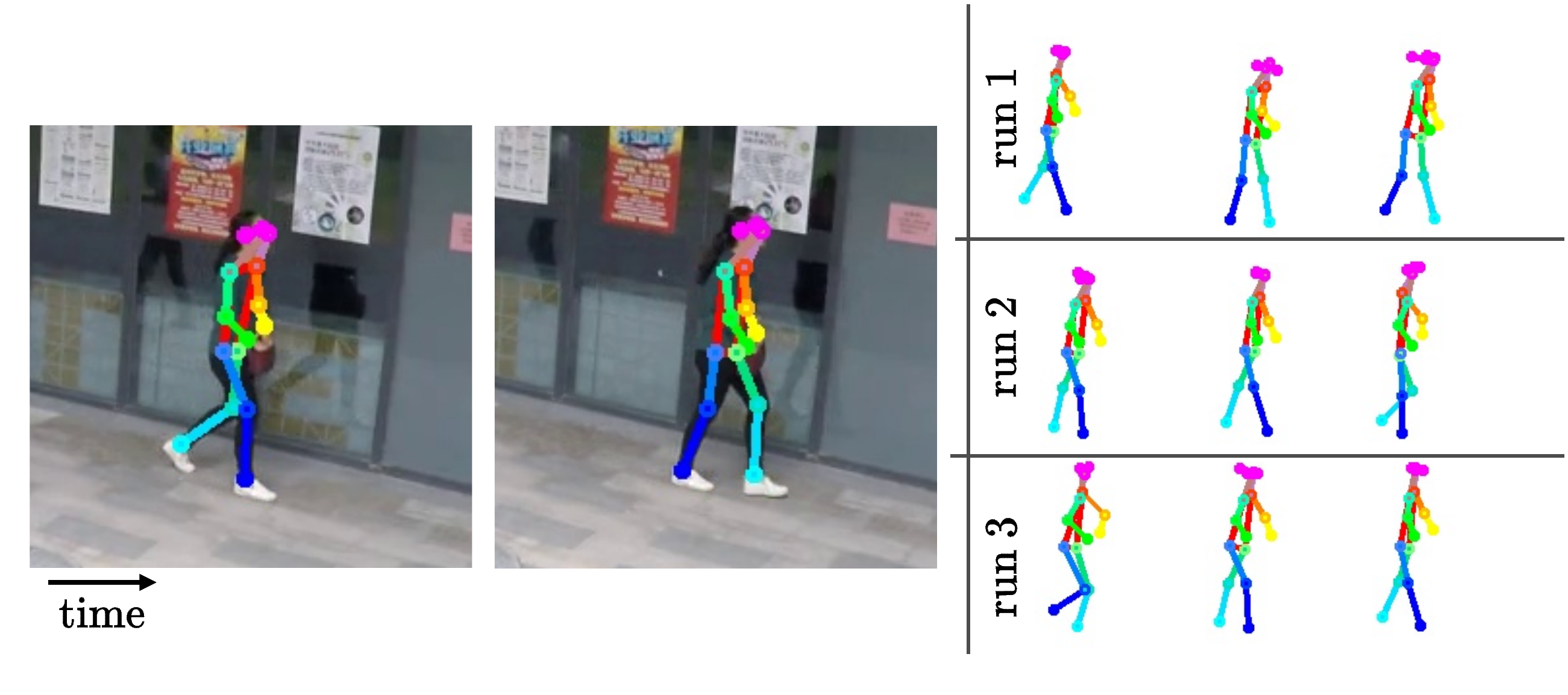}
    \caption{\textbf{Multiple \textcolor{future}{Ftr} on HR-STC}. Given the same input, we reconstruct future skeletons for each run corresponding to the model from Table~\ref{tab:multi_run}. Predicted future joints vary across runs however all models learn robust representations of expected normal behavior.}
    \vspace{-1em}
    \label{fig:multi_run}
\end{figure}

% Multi-run
\noindent
\textbf{Multiple predictions}. An important aspect of the reconstruction of normal behaviors corresponding to future, past, or present events is their predictability. VAD models need to robustly extrapolate (or interpolate) occluded trajectory segments of expected normal behaviors. In Table~\ref{tab:multi_run} we show AUC scores over multiple runs on HR-STC and HR-Avenue, without any notable changes in the performance being observed. Examples of predicting future keypoints for each run from Table~\ref{tab:multi_run} are shown in Figure~\ref{fig:multi_run}. Albeit their differences, all three of the reconstructed skeletons do correspond to expected normal motions.

%------------------------------------------------------------------------
\section{Conclusions}
\label{sec:conclusion}

We have proposed a holistic representation of trajectories with past, present, and future segments for VAD. Based on these segments, we introduce a multitask approach to capture normality in trajectories and jointly reconstruct past, present, and future segments. Temporally occluded trajectories are encoded and combined with a learned tensor. The latent representations of the occluded segments are learned through self-supervision with positive, soft-negative, and hard-negative pairs for each temporal location of the latent. Representations for the entire trajectory are decoded back to the input space and compared to the ground truth trajectory. Extensive experiments over three VAD datasets demonstrate the effectiveness of our approach. Additionally, we are the first to investigate the prediction of abnormal trajectories in the past as well as the present. We believe that this holistic representation of trajectories is a promising direction for future VAD research. 

\noindent
\textbf{Acknowledgments}. We use publicly available datasets. Research is funded by imec.icon Surv-AI-llance project and FWO (Grant G0A4720N).

%%%%%%%%% REFERENCES
{\small
\bibliographystyle{ieee_fullname}
\bibliography{egbib}
}

\clearpage

% Pages are numbered in submission mode, and unnumbered in camera-ready
\setcounter{section}{0}
\setcounter{equation}{0}
\setcounter{figure}{0}
\setcounter{table}{0}
\SupplementaryMaterial

\twocolumn[{%
\begin{center}
    \centering
    \textbf{\Large{Holistic Representation Learning for Multitask Trajectory Anomaly Detection \\ Supplementary Material}}\\
    \hspace{2em}
    \vspace{3em}
\end{center}%
}]

\begin{table}[t]
\centering
\caption{\textbf{Encoder hyperparameter optimization on HR-STC}. The setting with the best overall score is in \textcolor{lightgray}{light gray}. The best AUC score per task is in \textbf{bold}. We note that $\lambda_s$ (for joints) and $\lambda_{bb}$ (for box corners) are only used for trajectory reconstruction.}
\resizebox{\linewidth}{!}{%
\begin{tabular}{l l l l l c c c }
\hline
 lr & $\beta$ & $\gamma$ & $\lambda_{s}$ & $\lambda_{bb}$ & \textcolor{future}{\textbf{Ftr}} & \textcolor{present}{\textbf{Prs}} & \textcolor{past}{\textbf{Pst}}  \tstrut \bstrut \\
\hline
\multirow{2}{*}{$5e^{-4}$} & 0.001 & 0.1 & 2 & 4 & 75.6 & 72.8 & 74.7 \\
& 0.001 & 0.1 & 3 & 5 & 75.9 & 72.9 & 75.0 \\  
\hline
\multirow{6}{*}{$1e^{-4}$} & 1.0 & 0.1 & 3 & 5 & 76.9 & 73.1 & 75.1 \\ 
& 0.1 & 0.1 & 3 & 5 & 77.4 & 73.2 & 75.4 \\ 
& 0.01 & 0.1 & 3 & 5 & 77.7 & 73.4 & 75.8 \\
& 0.001 & 1.0 & 3 & 5 & 76.8 & 72.9 & 74.5 \\
& \cellcolor{LightGrey}0.001 & \cellcolor{LightGrey}0.1 & \cellcolor{LightGrey}3 & \cellcolor{LightGrey}5 & \cellcolor{LightGrey}\textbf{77.9} & \cellcolor{LightGrey}73.5 & \cellcolor{LightGrey}\textbf{75.7} \\
& 0.001 & 0.01 & 3 & 5 & 77.7 & \textbf{73.6} & 75.4 \\
\end{tabular}
}
\label{tab:encoder_hp}
\end{table}

\begin{table}[t]
\centering
\caption{\textbf{HR-STC AUC scores over different} $\lambda$ \textbf{hyoperparameters for \textcolor{future}{Ftr}}. Best combination of $\lambda_{s}$ and $\lambda_{bb}$ is in \textcolor{green}{green}. Settings for which their AUC is above the state-of-the-art are in \textbf{bold}. Settings that were not explored are denoted with N/A in \textcolor{lightgray}{gray}.}
\resizebox{\linewidth}{!}{%
\begin{tabular}{l | c c c c c c c c c}
\hline
 & \multicolumn{9}{c}{$\lambda_{s}$} \tstrut \bstrut \\
\cline{1-10}
\multirow{9}{*}{$\lambda_{bb}$} & & 1 & 2 & 3 & 4 & 5 & 6 & 7 & 8 \tstrut \bstrut \\ 
\cline{3-10}
& \multicolumn{1}{c|}{1} & 76.5 & \textcolor{lightgray}{N/A} & \textcolor{lightgray}{N/A} & \textcolor{lightgray}{N/A}  & 76.2 & \textcolor{lightgray}{N/A}  & 75.6 & 75.9 \tstrut \\
& \multicolumn{1}{c|}{2} & \textcolor{lightgray}{N/A}  & 75.9 & 77.3 & \textbf{77.6} & 77.3 & \textcolor{lightgray}{N/A}  & 76.9 & 76.7 \\
& \multicolumn{1}{c|}{3} & 76.5 & 76.8 & \textcolor{lightgray}{N/A} & 77.5 & \textcolor{green}{77.9} & \textbf{77.6} & 77.2 & 76.9 \\
& \multicolumn{1}{c|}{4} & \textcolor{lightgray}{N/A} & \textcolor{lightgray}{N/A}  & \textcolor{lightgray}{N/A}  & 76.3 & 76.5 & 76.9 & 77.4 & 77.2 \\
& \multicolumn{1}{c|}{5} & \textcolor{lightgray}{N/A}  & \textcolor{lightgray}{N/A} & 76.4 & 76.9 & 76.6 & 76.4 & 76.7 & 76.7 \\
& \multicolumn{1}{c|}{6} & 76.1 & 76.3 & 76.4 & 76.6 & 76.7 & 76.4 & 76.2 & N/A \\
& \multicolumn{1}{c|}{7} & 76.4 & 76.2 & 76.2 & 76.5 & 76.8 & 76.8 & 77.0 & 76.9 \\
& \multicolumn{1}{c|}{8} & \textcolor{lightgray}{N/A} & \textcolor{lightgray}{N/A} & \textcolor{lightgray}{N/A} & 76.3 & \textcolor{lightgray}{N/A} & \textcolor{lightgray}{N/A}  & \textcolor{lightgray}{N/A}  & \textcolor{lightgray}{N/A}  \\
\end{tabular}
}
\label{tab:lambdas}
\vspace{-1.1em}
\end{table}

%%%%%%%%% BODY TEXT
\section{Encoder and decoder hyperparameters}
\label{sec:hyperparams}

We discover the optimal hyperparameters with random search. For the encoder hyperameters we define a search space of $\beta \in \{ 1.0,0.1,0.01,0.001\}$ and $\gamma \in \{ 1.0, 0.1, 0.01 \}$ where $\beta$ is the hyperparameter used by the regularizer for soft negative pairs $S^-$ and $\gamma$ is the margin hyperparameter used in the triplet loss. A full list of searched combinations is shown in Table~\ref{tab:encoder_hp}. The largest decreases in performance are observed for $\beta=1.0$ corresponding to strong penalization for $S^-$ pairs. As we also motivate in Section~\ref{sec:experiments::results}, adjacent points from the same trajectory are bound to include similarities therefore strong penalization pushes representations of adjacent points further apart. 

We additionally perform a random search over $\lambda$ decoder hyperparameters for skeleton joints, denoted as $\lambda_s$, and bounding box corners, denoted as $\lambda_{bb}$. As shown in Table~\ref{tab:lambdas}, the general trend for the settings with AUC scores comparable to those of the state-of-the-art is $\lambda_{bb} < \lambda_s$ and $\lambda_{bb} < 5$. We believe that $\lambda_{bb} < \lambda_s$ is due to the trajectories of bounding box corners significantly varying based on the orientation of the pose being tracked. In contrast, the trajectories of spatial locations of skeleton joints are constrained to the corresponding body parts.

\section{Additional predicted segment examples}
\label{sec:examples}

Our proposed multitask approach for anomaly detection is based on the holistic representation of trajectories as segments. Crucially, the continuity of trajectories is based on the detection of skeletons at the pose level. Skeletons or individual keypoints not detected by the pose detector would correspond to partial ground truth trajectories for which their reconstructions are not possible. As with the ablation results in Section~\ref{sec:experiments::qualitative} of the main text, we provide qualitative results for \textcolor{future}{Ftr}, \textcolor{present}{Prs}, and \textcolor{past}{Pst} extrapolation and interpolation tasks.

\begin{figure*}[t]
    \centering
    \begin{subfigure}[b]{\textwidth}
        \includegraphics[width=\linewidth]{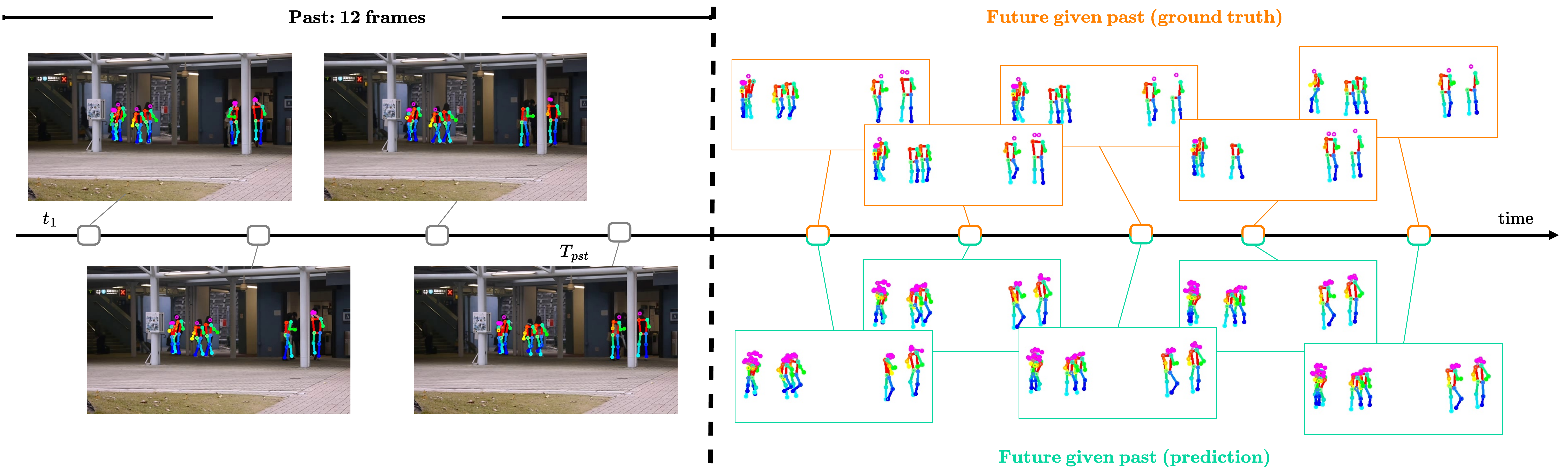}
        \caption{\textbf{Predicting the future given the past} (\textcolor{future}{Ftr}).}
    \label{fig:examples:extrapolation_future_normal_sp}
    \end{subfigure}
    \begin{subfigure}[b]{\textwidth}
        \vspace{.5em}
        \includegraphics[width=\linewidth]{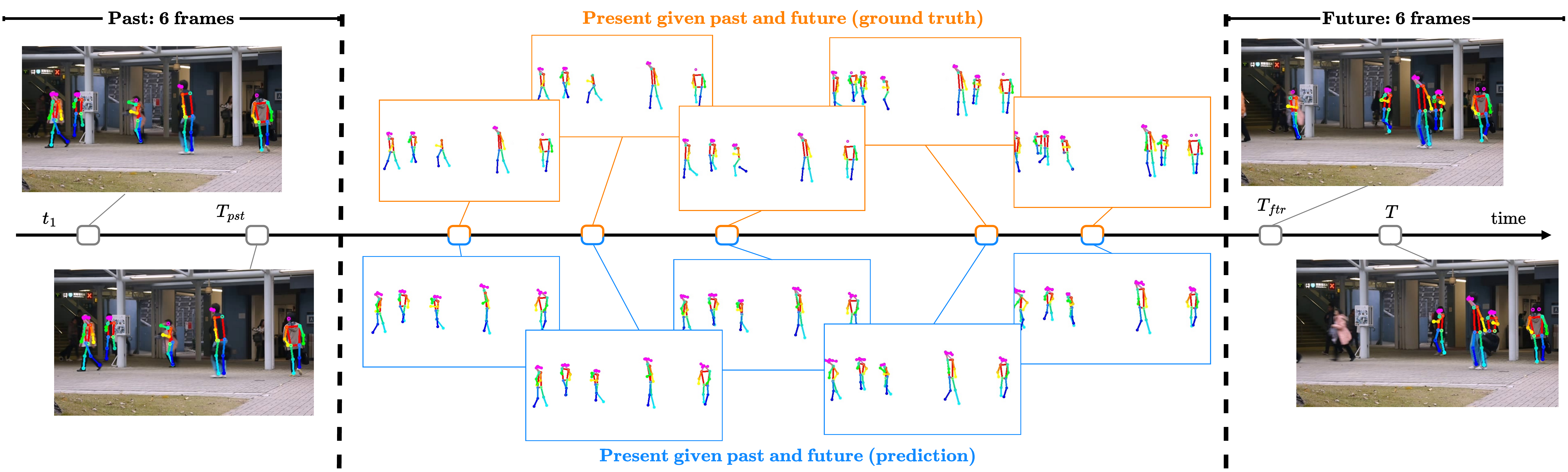}
        \caption{\textbf{Predicting the present given both the past and future} (\textcolor{present}{Prs}).}
    \label{fig:examples:interpolation_normal_sp}
    \end{subfigure}
    \begin{subfigure}[b]{\textwidth}
        \vspace{.5em}
        \includegraphics[width=\linewidth]{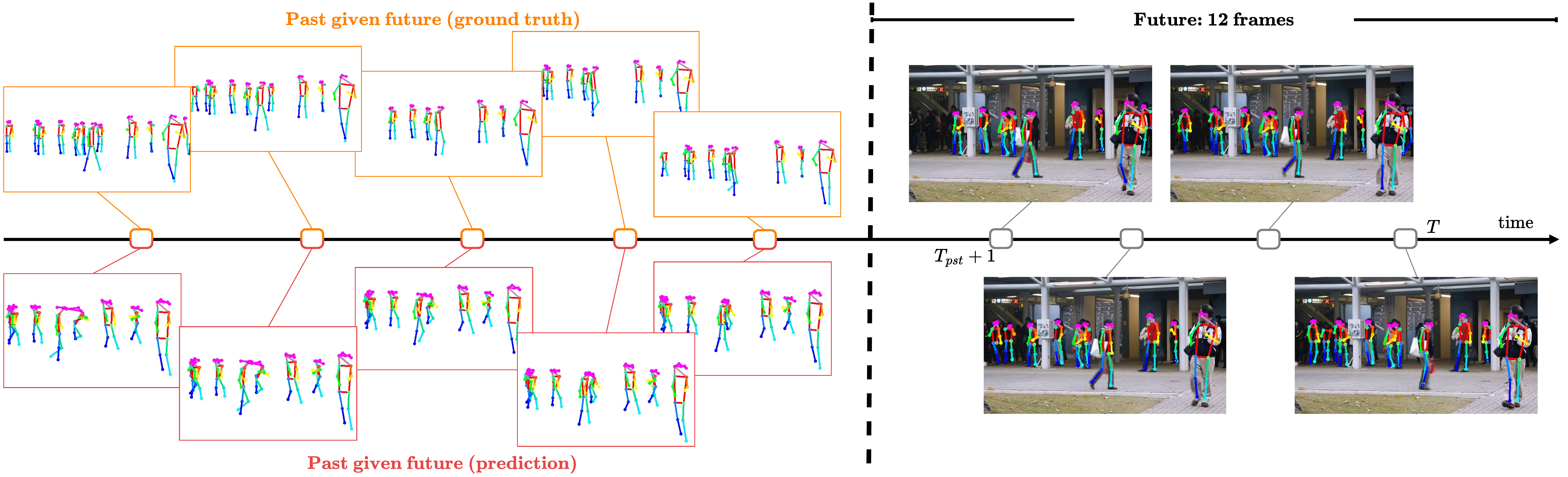}
        \caption{\textbf{Predicting the past given the future} (\textcolor{past}{Pst}).}
    \label{fig:examples:extrapolation_past_normal_sp}
    \end{subfigure}
    \caption{\textbf{Skeleton sequence reconstruction in HR-Avenue}. Input sequences are keypoints from 18 frames with each of the predicted segments for \textcolor{future}{Ftr}, \textcolor{present}{Prs}, and \textcolor{past}{Pst} task being 6 frames. In each of the occluded segments, not all keypoints are recognized by the pose detector. This relates to partial ground truths. As shown the model can learn strong representations with predicted keypoints still being sensible despite not being detected and part of the ground truth.}
    \label{fig:examples_sp}
    \vspace{-1em}
\end{figure*}

Figure~\ref{fig:examples_sp} shows examples of non-continuous trajectories and reconstructions. Across all tasks, our model is capable of reconstructing continuous trajectories despite the absence of undetected keypoints in the ground truth. For either extrapolation tasks \textcolor{future}{Ftr} and \textcolor{past}{Pst} partially observed trajectories are reconstructed for the entire skeletons regardless of the keypoints detected. For cases in which entire skeletons are not detected, reconstructions are also not created. Specifically, the leftmost skeletons in the \textcolor{past}{Pst} example are not detected in the majority of the observable frames and in turn, cannot be reconstructed/predicted. In contrast, predictions for absent keypoints, as shown in the \textcolor{future}{Ftr} example, can still be made as the model learns expected motions with respect to the remaining observable keypoints in the trajectory. In example of the \textcolor{present}{Prs} interpolation task, reconstructions are only done for skeletons that are correctly detected in both proceeding or succeeding frames. As shown, predictions given learned motions are also made for the joints that are not observable.

\begin{table}[t]
\centering
\caption{\textbf{AUC scores on HR-STC and HR-Avenue over varying occluded segment lengths for training and inference}. The default settings in which the occluded segment lengths are the same for both training and inference are denoted in \textcolor{lightgray}{gray}.}
\resizebox{\linewidth}{!}{%
\begin{tabular}{l l c c c c c c c }
\hline
\multirow{2}{*}{$\uparrow \! |\widehat{\mathbf{t}}|$}&
\multirow{2}{*}{$\downarrow \! |\widehat{\mathbf{t}}|$}&
\multicolumn{3}{c}{\textbf{HR-STC}} &&
\multicolumn{3}{c}{\textbf{HR-Avenue}} 
\tstrut \bstrut \\
\cline{3-5} \cline{7-9}
\tstrut & & \textcolor{future}{\textbf{Ftr}} & \textcolor{present}{\textbf{Prs}} & \textcolor{past}{\textbf{Pst}} && \textcolor{future}{\textbf{Ftr}} & \textcolor{present}{\textbf{Prs}} & \textcolor{past}{\textbf{Pst}} \bstrut \\
\hline
\multirow{2}{*}{6} & \cellcolor{LightGrey}6 & \cellcolor{LightGrey}77.9 & \cellcolor{LightGrey}73.5 & \cellcolor{LightGrey}75.7 &\cellcolor{LightGrey}& \cellcolor{LightGrey}89.4 & \cellcolor{LightGrey}86.3 & \cellcolor{LightGrey}87.6 \\
& 12 & 72.3 & 69.6 & 69.8 && 85.7 & 83.2 & 84.4 \\
\hline
\multirow{2}{*}{12} & 6 & 72.3 & 70.6 & 72.1 && 85.0 & 81.3 & 84.7 \\
& \cellcolor{LightGrey}12 & \cellcolor{LightGrey}75.9 & \cellcolor{LightGrey}72.7 & \cellcolor{LightGrey}75.9 &\cellcolor{LightGrey}& \cellcolor{LightGrey}87.9 & \cellcolor{LightGrey}85.5 & \cellcolor{LightGrey}86.4 \\
\end{tabular}
}
\label{tab:seq_len_sp}
\end{table}

\section{Ablations on reconstruction lengths}
\label{sec:lengths}

Motivated by Table~\ref{tab:seq_len} in the main paper, we further 
 jointly ablate over occluded segment lengths for training and inference in Table~\ref{tab:seq_len_sp}. Normally, the model is optimized to extrapolate/interpolate occluded segments of predefined length. However, scenarios may exist in which retraining or finetuning for specific lengths may not be available. We thus train our model with occluded segment lengths of either $|\widehat{\textbf{t}}|=6$ or $|\widehat{\textbf{t}}|=12$. In each setting, the encoder learns to pull closer latent and learned representations for the occluded segments while the decoder learns to reconstruct the full trajectory of $|\textbf{t}|=18$. For each setting, we run inference with either $|\widehat{\textbf{t}}|=6$ or $|\widehat{\textbf{t}}|=12$ occluded segments. For clarity, training and inference lengths are denoted with $\uparrow\!|\widehat{\textbf{t}}|$ and $\downarrow\!|\widehat{\textbf{t}}|$ respectively. On average, a
-3.1 and -2.6 decrease in the AUC score is observed over both HR-STC and HR-Avenue on the $\uparrow\!|\widehat{\textbf{t}}|=12$ setting when changing the inference occlusion from $\downarrow\!|\widehat{\textbf{t}}|=12$ to $\downarrow\!|\widehat{\textbf{t}}|=6$. A similar drop is observed in the $\uparrow\!|\widehat{\textbf{t}}|=6$ setting when changing $\downarrow\!|\widehat{\textbf{t}}|=6$ to $\downarrow\!|\widehat{\textbf{t}}|=12$ with -5.1 and -3.3 AUC score reductions for HR-STC and HR-Avenue. This demonstrates that despite not being optimized during training, sensible results can still be obtained in inference cases where occluded segments are of varying lengths and finetuning is not possible.

\end{document}